# Generative Design of Crystal Structures by Point Cloud Representations and Diffusion Model

## Author Information

[1,2]Zhelin Li (https://orcid.org/0009-0000-7050-312X), [1,2]Rami Mrad (https://orcid.org/0009-0002-8350-0141), [1,2]Runxian Jiao (https://orcid.org/0009-0006-3852-6739) , [1,2]Guan Huang, [1,2]Jun Shan,*, [1,2]Shibing Chu (https://orcid.org/0009-0000-8909-1786)& *,[1,2]Yuanping Chen(https://orcid.org/0000-0001-5349-3484)

## Highlight

We present a materials generation framework for bulk materials based on diffusion model.

We use point cloud presentation to encode atomic position information.

We generate batches of crystals and many of them have the potential for stability.

We validate our results with first-principles calculations

## Abstract

[1] School of Physics and Electronic Engineering, Jiangsu University, Zhenjiang, Jiangsu 212013, PR China

[2] Jiangsu Engineering Research Center on Quantum Perception and Intelligent Detection of Agricultural Information, Zhenjiang, 212013, China

*Correspondence:

c@ujs.edu.cn (S.C.),

chenyp@ujs.edu.cn (Y. C.)

Efficiently generating energetically stable crystal structures has long been a challenge in material design, primarily due to the immense arrangement of atoms in a crystal lattice. To facilitate the discovery of stable materials, we present a framework for the generation of synthesizable materials leveraging a point cloud representation to encode intricate structural information. At the heart of this framework lies the introduction of a diffusion model as its foundational pillar. To gauge the efficacy of our approach, we employed it to reconstruct input structures from our training datasets, rigorously validating its high reconstruction performance. Furthermore, we demonstrate the profound potential of point cloud-based crystal diffusion (PCCD) by generating entirely new materials, emphasizing their synthesizability. Our research stands as a noteworthy contribution to the advancement of materials design and synthesis through the cutting-edge avenue of generative design instead of conventional substitution or experience-based discovery.

## Introduction

The continuous advancement of technology hinges significantly on the development of new materials, making it essential to unravel the complex relationships between molecular or crystal structures and their properties. Currently, two main methods are used for designing crystal structures: altering existing materials using scientific intuition and empirical principles or global optimization algorithms[1] and mining material databases with the Materials Project (MP)[2], known as high-throughput virtual screening[3], which has shown great success in various applications. However, the computational expense associated with density functional theory (DFT) calculations renders an exhaustive search of the theoretical material space infeasible[4]. In recent years, there has been a notable surge in research dedicated to harnessing artificial intelligence for the exploration of new materials[5-10]. However, within the field of crystallography, the predominant application of ML techniques is focused on predicting material properties, such as composition, band gap, or formation energy[11-13]. Consequently, the utilization of ML algorithms for crystal generation remains relatively nascent, underscoring the pressing need for the further development of artificial intelligence-generated content (AIGC) within the realm of crystallography.

In the field of material exploration, generative models have been proven to be particularly effective[7]. Over the past few years, two fundamental models have been widely applied: the generative adversarial network (GAN)[14] and the variational autoencoder (VAE)[15]. Currently, an array of studies has been dedicated to structure generation, drawing on the capabilities of these two models. An example is the study conducted by Jordan Hoffmann *et al.*[16], in which voxel representation was employed for crystals, a VAE was utilized for voxel data generation, and a U-Net model was subsequently applied for voxel classification. Zekun Ren *et al.*[5] employed VAE for the reverse design of materials. Kim *et al.*[7] utilized a GAN model to explore structures within the Mg-Mn-O ternary system, while Baekjun Kim *et al.*[8] employed a Wasserstein generative adversarial network (WGAN) in their quest to discover crystalline porous materials. These research endeavors highlight the versatility and promise of generative models in the context of material discovery and design.

Recently, there has been a significant emergence of models for generating generic crystal structures. A notable example is the Crystal Diffusion Variational Autoencoder (CDVAE) developed by Tian Xie et al.[6], which successfully integrates a diffusion model with a Variational

Autoencoder (VAE) for crystal generation. Furthermore, the Cond-CDVAE model[17] extends this approach by allowing the incorporation of user-defined material and physical parameters, such as composition and pressure. Another major breakthrough in this domain is MatterGen[18], which is capable of generating stable and novel materials with specified chemical compositions, symmetries, and mechanical, electronic, and magnetic properties.

Nevertheless, most models address the challenge of how to improve the quality of generation results[19]. Jonathan Ho *et al*.[20] introduced a novel generative model known as the denoising diffusion probabilistic model (DDPM). Notably, various research teams, such as OpenAI[21,22], NVIDIA[23] and Google[24], have achieved significant breakthroughs in the application[25] of this model. Considering its excellent generative capability, we aim to investigate the latent potential of this model in the domain of structure generation and its potential to enhance the creative aspects of the model. Additionally, to minimize computational expenses and tailor diffusion modeling, we propose a point cloud representation[26] to encode atom sites, element information, and lattice constants.

In this paper, we introduce a streamlined deep learning framework for crystal generation: point cloud-based crystal diffusion (PCCD). To test the model's reliability, we intentionally added noise to our dataset and then used the PCCD to reconstruct the majority of the inputs with only minor deviations. Furthermore, we calculated the energy above hull ($E_{hull}$) per atom for a set of crystal structures generated by PCCD, revealing that many of these structures had low energy values, indicating their potential significance. Furthermore, our analysis revealed structures not in the database or with a stable phonon structure, emphasizing the ability of

PCCD to generate new and potentially significant crystal structures.

## Results and Discussion

In the PCCD, the training of the diffusion model involves the incremental addition of noise, with the model essentially learning how to peel noise from the corrupted data. In an ideal scenario, saving the data from the database, along with the added noise, should enable the eventual reconstruction of these original data without noise. To validate the model's effectiveness, we selected a batch of structures from the database as the test set and performed 1000 iterations of noise to obtain and store the results. These noise-augmented results were then used as inputs for the PCCD instead of true random numbers. In theory, 1000 times of noise should be removed, and the data should be restored. To ensure accurate atomic site matching, the statistics presented are based on 868 samples, as only structures with matching atom counts can be compared. For the purpose of predicting atomic coordinates, we align each atom in the predicted crystal structure with its counterpart in the original crystal, given that both structures have the same total number of atoms. The distance between each atom in one structure and each atom in the other structure is calculated, taking into account translational symmetry. This symmetry allows atoms in the original crystal to be matched with atoms in adjacent cells of the predicted crystal, effectively aligning coordinates such as (0,0,0) in one structure with (1,1,1) in the other when the distance is zero. Finally, the Greedy Algorithm is employed to perform the matching after all distances have been determined. We then compared the restored data to the original dataset, as illustrated in Fig. 1, providing a robust assessment of the model's reconstruction performance. This experiment serves as a rigorous validation of the model's capabilities.

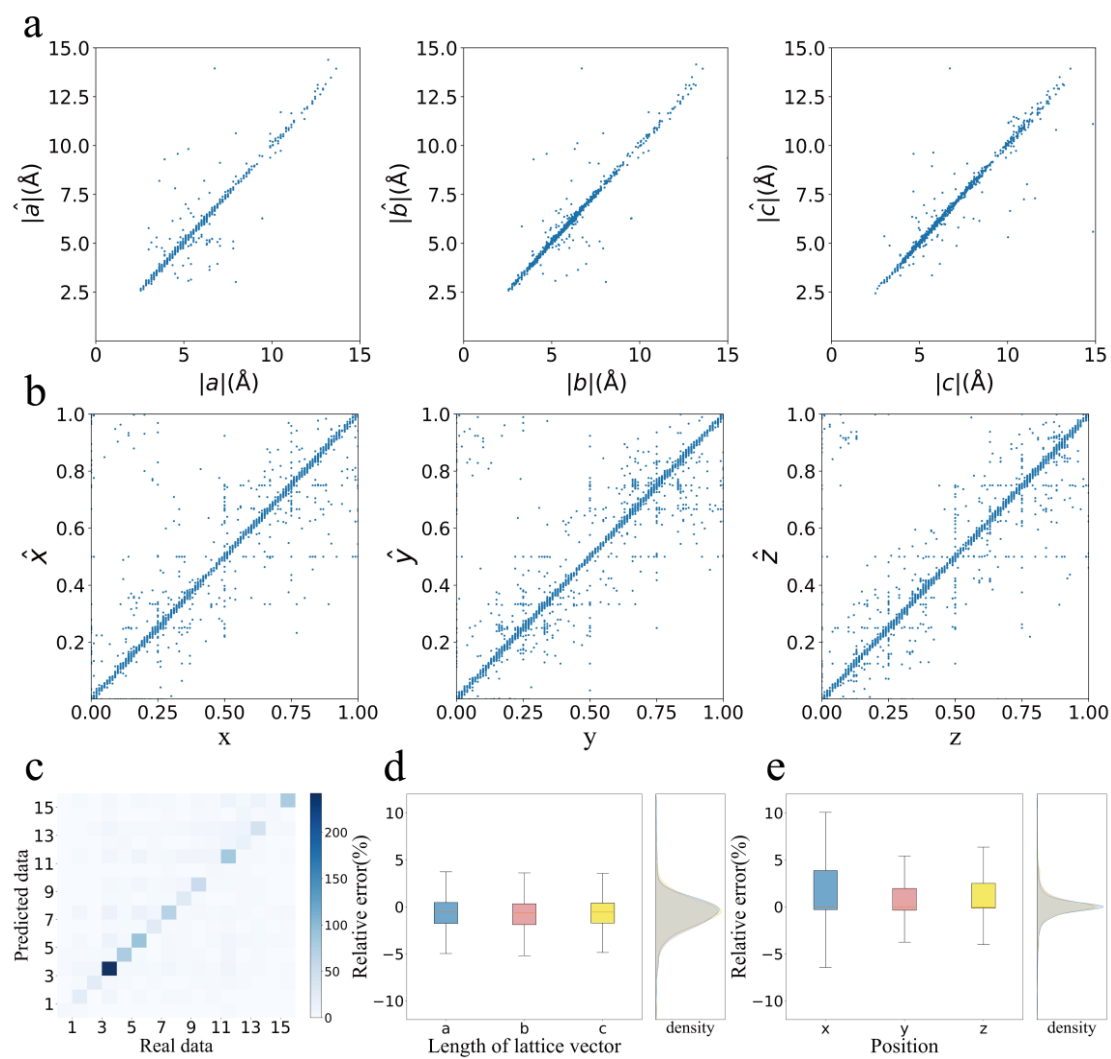

**Fig. 1|Reconstruction results of the PCCD. a.** The parity plots for lattice lengths of reconstructed materials and original materials. **b.** The parity plots for the atomic positions of the reconstructed materials and original materials. **c.** Heatmap of the atom number relationship between the reconstructed materials and original materials. **d.** Box plot of the lattice length relative error with density distribution. **e.** Box plot of the relative errors of the atomic x, y, and z coordinates with respect to the density distribution.

Given that we did not specify the number of atoms in the PCCD, the accuracy of predicting the atom count serves as a direct indicator of the model's performance. In Fig. 1c, a heatmap illustrates the relationship between the sum of atoms in the original data and the data predicted by the PCCD. Notably, a clear diagonal line represents accurate predictions, and the accuracy rate is 67.81% (868 out of 1280 samples).

Among these 868 samples with accurately predicted atom counts, we further calculated the relative errors for the lattice length of each corresponding atom, where $|a|, |b|, |c|$ are the original lattice lengths and $|\hat{a}|, |\hat{b}|, |\hat{c}|$ are the predicted lattice lengths (as shown in Fig. 1a) and their coordinates (depicted in Fig. 1b). A significant portion of these errors is visibly clustered around the $y = x$ line. To gain deeper insights into their distribution, we conducted a detailed analysis using box plots for both aspects, as presented in Fig. 1d and Fig. 1e. Fig. 1d displays the box plot for the relative errors of the lattice parameters $a$, $b$, and $c$. The acceptable ranges for $a$, $b$, and $c$ typically fall within upper limits of 3.81%, 3.66%, and 3.57%, respectively, and lower limits of -5.12%, -5.23%, and -4.90%, respectively. The effective rates for these parameters are as follows: 88.48% for lattice parameters $a$, 90.55% for lattice parameters $b$, and 89.40% for lattice parameters $c$. Alongside the box plots, the kernel density function plots help illustrate the concentration of the data. In Fig. 1e, we present the box plot for the relative errors of the $x$, $y$, and $z$ coordinates for each atom. The typical acceptable ranges for $x$, $y$, and $z$ coordinates fall within the upper limits of 10.05%, 5.44%, and 6.44%, respectively, and lower limits of -6.50%, -3.84%, and -4.09%, respectively. The effective rates for these coordinates are calculated as 70.83% for $x$, 71.60% for $y$, and 72.02% for $z$. Most of the errors are relatively small, and they can be readily corrected during DFT geometry optimization. This analysis indicates that the framework is already functional and effective.

However, we also conducted a more in-depth analysis to explore the objective factors that may influence model errors.

|  | *a* | *b* | *c* | *x* | *y* | *z* |
|---|---|---|---|---|---|---|
| Efficiency | 88.48% | 90.55% | 89.40% | 70.83% | 71.60% | 72.02% |
| Upper limits | 3.81% | 3.66% | 3.57% | 10.05% | 5.44% | 6.44% |
| Lower limits | -5.12% | -5.23% | -4.90% | -6.50% | -3.84% | -4.09% |

**Table 1|Data details for Fig. 1.** Efficiency calculations according to the box plots in Fig. 1d and Fig. 1e. The upper and lower limits are shown for the box plots in Fig. 1d and Fig. 1e, respectively.

A significant contributing factor to the performance limitations of PCCD lies in the preprocessing stage before training. To facilitate the normalization of lattice vectors and enhance reversibility, all lattice vector data (data in the 3rd channel) were divided by 15. Consequently, the model's capacity was restricted to generating values within the range of -1 to 1, which, in turn, led to a limitation in predicting lattice vectors. Specifically, the model could predict only lattice vectors with a maximum length of 15 Å. As a consequence, structures featuring lattice vectors exceeding the maximum could not be accurately predicted. Upon calculating the relative errors for the lattice lengths of each atom across all samples (a total of 1280), we observed less favorable outcomes due to this limitation. The box plots vividly illustrate that the effective rates for lattice vector length were only 73.98% for the '$a$' lattice parameter, 75.39% for '$b$', and 71.64% for '$c$'. This highlights the significant correspondence between the errors in both the atom counts and lattice vector length predictions.

Notably, many of these errors were associated with structures featuring at least one lattice vector longer than the maximum. This suggests a substantial interrelation between the atom count and lattice vector predictions, despite their presence in different data channels. Furthermore, the comparison of the $x, y$ and $z$ positions of atoms, as shown in the three figures in Fig. 1b, reveals that a portion of the data clustered around the position (0,1). These data points were excluded when calculating statistics, as they were considered erroneous. However, it is important to note that crystal cells are periodic, and such data points are essentially equivalent to (0,0) or (1,1). This periodicity factor contributes to lower accuracy in the statistical analysis.

In order to describe the matching relationships between structures more accurately, we selected several statistical metrics that are suitable for Crystal Structure Prediction (CSP)[27]. For each pair of crystals composed of a reconstructed structure and its original structure, we calculated their Energy distance (Fig. 2a), Orbital Field Matrix distance (Fig. 2b), CrystalNN Fingerprint distance (Fig. 2c), Superpose distance (Fig. 2d), RMS Anonymous distance (Fig. 2e) and Graph Edit distance (Fig. 2f). Box plots and kernel density function graphs can reflect their distribution situation. Fig. 2 presents the outputs of the statistics, from which we can deduce that, for the energy distance, the reconstruction process of the model consists of obtaining values close to 0, proving the efficiency of the crystal reconstruction process. The box plot in Fig. 2a shows that the largest portion of the reconstructed data are near to 0. Fig. 2b illustrates the Orbital Field Matrix distance, where the density of the reconstructed data shows a peak at approximately 0; here, we can conclude that the reconstruction process is efficient. The CrystalNN Fingerprint distance is a machine learning-based approach that investigates the number of neighboring atoms in the same or similar crystal structures; hence, Fig. 2c shows the repartition of the data, which proves the similarity of the reconstructed data. Fig. 2d shows the superpose distance, which reflects the efficacy of the model in the training process, where these metrics are used to compare the similarity of the periodic structure. Illustrating the RMS anonymous distance in Fig. 2e is an absolute shoring to our model where we constate the similarity of the reconstructed data in these evaluation metrics and prove the model capacity. By exploring Fig. 2f and the last performance metrics that we chose for assessing our work, the Graph Edit distance, in this case, we can deduce also deduce the success of the reconstruction process due to the output of this metric, where we compare the number of edges and nodes; hence, we notice the peak at 0 to prove the structural similarity in terms of reconstruction.

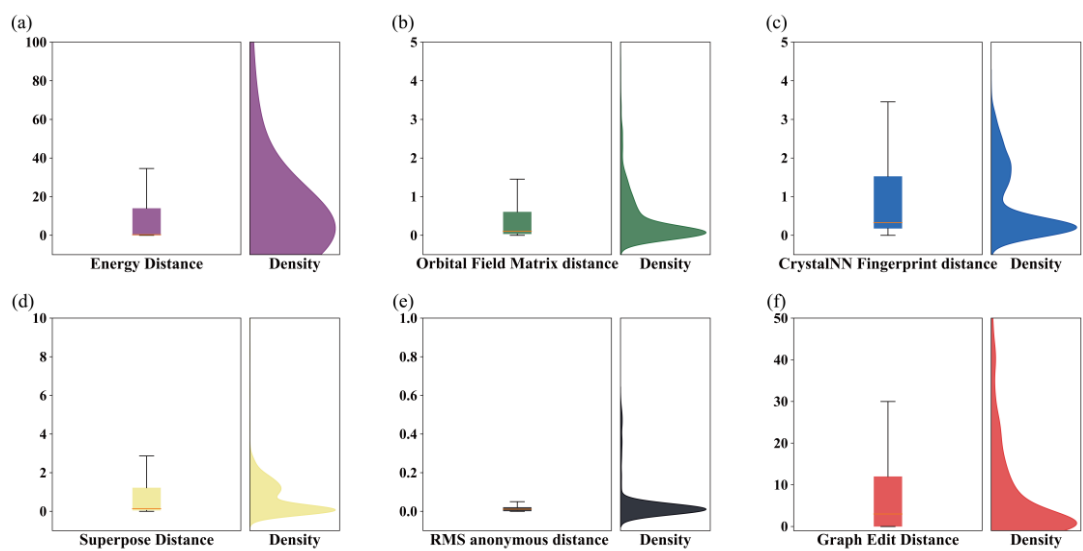

**Fig. 2| Statistics for matched structures.** Box plots and kernel density functions of the Energy distance (a), Orbital Field Matrix distance (b), CrystalNN Fingerprint distance (c), Superpose distance (d), RMS Anonymous distance (e) and Graph Edit distance (f) for

matched crystals.

In comparison to models tailored for specific material components, such as the Mg-Mn-O or $V_xO_y$ systems [7,28], PCCD demonstrates superior generalization capabilities. This means that we can effectively generate crystal structures composed of any combination of elements, provided that the total number of elements does not exceed three. As depicted in Fig. 3, PCCD enables the generation of unary systems (Fig. 3a), binary systems (Fig. 3b), and ternary systems (Fig. 3c), demonstrating its versatility and broad applicability. These findings also prove the diversity of this framework (Fig. 3d, Fig. 3e and Fig. 3f).

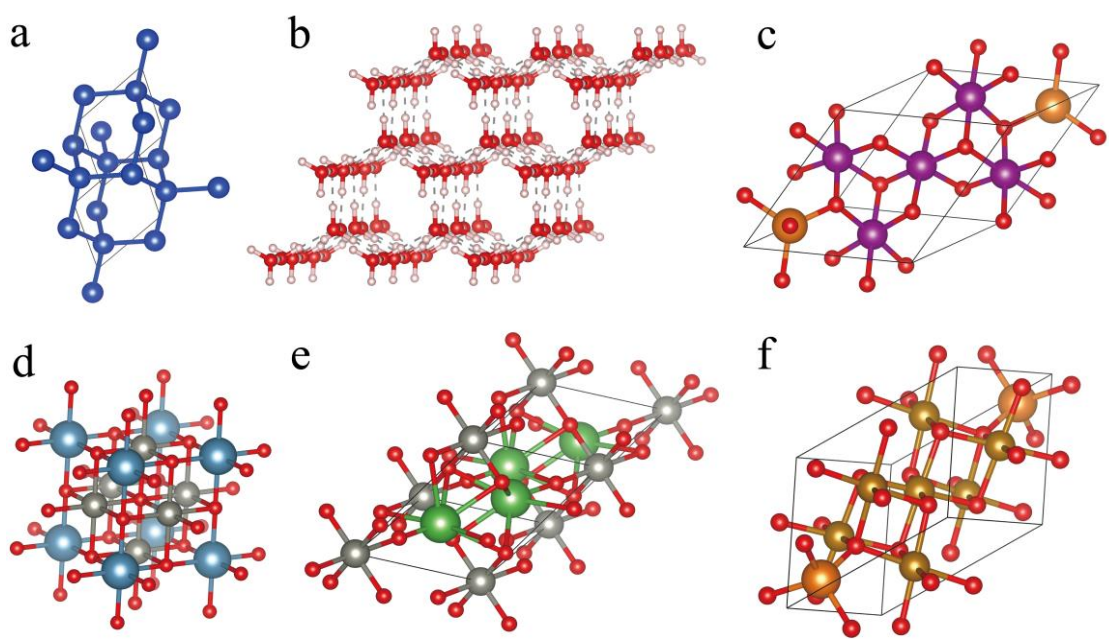

**Fig. 3|Examples of generated crystals. a.** Sample of the predicted data Si system. **b.** A sample of the predicted data 3 × 3 supercell for the H-O system ($H_2O$). **c.** Predicted data of the unit cell for the Mg-Mn-O system ($Mg_2Mn_3O_8$). **d.** Unit cell generated for $CaZn_3O_4$. **e.** Unit cell generated for $La_2ZnO_4$. **f.** Unit cell generated for $MgFe_2O_4$.

For statistical validation and comparative analysis with other models, we generated three distinct batches of structures. The first batch (batch #1) consists of all possible elements excluding noble gases and radioactive elements. The second batch (batch #2) includes rare earth elements, alkaline earth elements, transition metals, and oxygen selected due to their propensity to exhibit unique properties. The third batch (batch #3) comprises only commonly used elements. Following an initial screening process, we identified 1809, 746, and 120 structures in each batch, respectively. Of these, 1680 (92.87%), 669 (89.68%), and 108 (90%) structures were identified as novel discoveries.

We employed the Vienna Ab Initio Simulation Package (VASP)[29,30] to calculate the total energy. The generalized gradient approximation (GGA)[31] given by the Perdew–Burke–Ernzehof (PBE) parametrization[32] was used to describe exchange–correlation interactions. Furthermore, we utilized the *pymatgen* package to calculate the $E_{hull}$ per atom[33] and found that approximately 39.44%, 61.80% and 66.7% separately of the structures exhibited values less than 0.25 eV/atom (Fig. 4 and Tabel 2). In comparison, according to Yong Zhao's PGCGM[4], out of 1579 structures, 106 had values less than 0.25 eV/atom (5.3%). Sungwon Kim's model[7] and Juhwan Noh's model, known as iMatGen[28,] are two earlier models that have also made significant contributions to the field of GAN and VAE. Both of these authors assert that a structure with an $E_{hull}$ less than 80 meV/atom can be considered relatively stable. In their respective paper, Sungwon Kim's work obtained 113 results with an $E_{hull}$ per atom less than 80 meV/atom from 6000 generated structures, while iMatGen achieved 40 such results from 10,981 structures, with a ratio of 1.8% and 0.36%, respectively. In contrast, we identified 160,122 and 41 structures that met these criteria from generated structures in these three batches (8.9%, 16.35%, 34.17%).

It is important to note that in batch #2, the definition and calculation method of $E_{hull}$ may result in statistically anomalous values, either falsely high or low, particularly due to the limited representation of structures containing rare earth elements in the database. Consequently, we excluded structures containing rare earth elements from batch #2 and recalculated the statistics, as presented in Table 2. Further details can be found in the Supplemental Information. These findings suggest that the diffusion model may, to some extent, outperform GANs or VAEs in this field. It is worth highlighting that despite being a simplified version designed to explore the potential of the diffusion model and point cloud usage in the field of materials, our framework, akin to a pretrained model, has demonstrated comparable or even superior effectiveness in various aspects when compared to many other existing models.

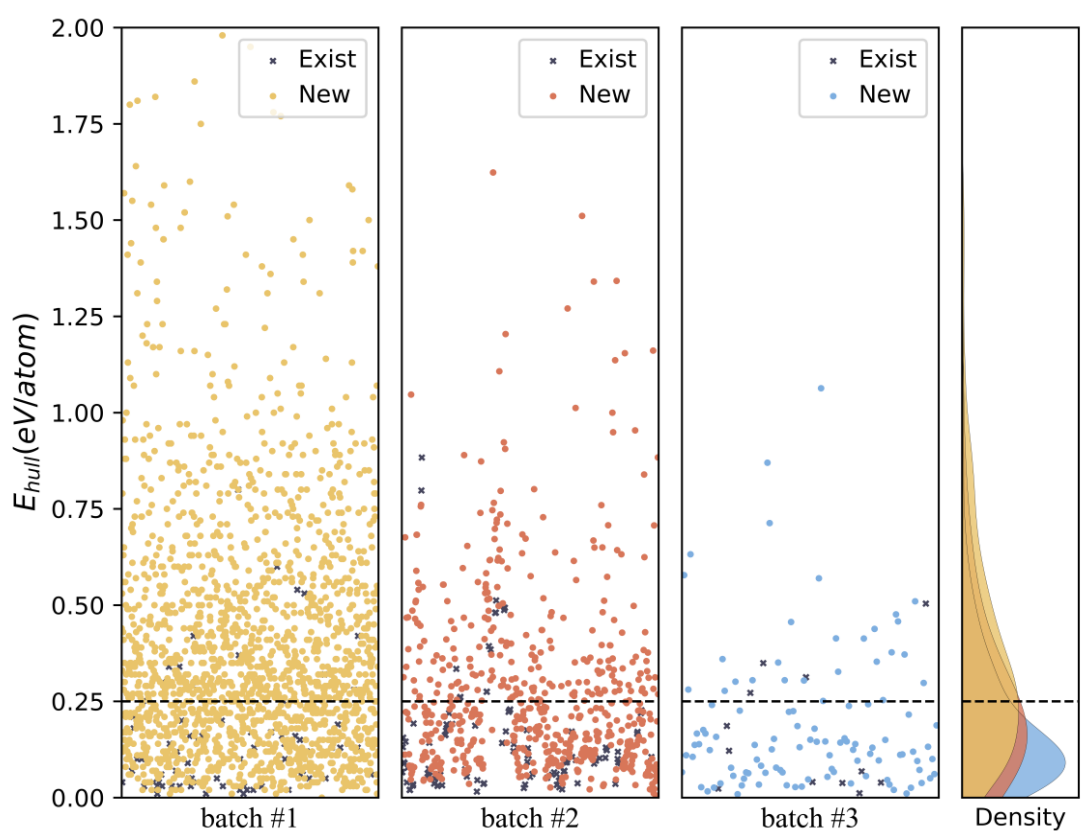

Fig. 4|**The distribution of $E_{hull}$ per atom for the generated data (three batches).** The

label "Exist" means the generated structure is present in the database, while the label "New" indicates that it is being reported for the first time. Batch #1 contains structures with any components. Batch #2 contains structures that consist of rare earth elements, alkaline earth elements, transition metal elements, and oxygen. Batch #3 contains structures that consisted of common elements.

| $E_{hull}$ | Batch #1 | Batch #2 | Batch #2 (without rare earth elements) | Batch #3 | PGCGM[4] | Sungwon Kim's model[7] | iMatGen[28] |
|---|---|---|---|---|---|---|---|
| 0.25eV/atom | 39.44% | 61.80% | 89.66% | 66.67% | 5.3%[4] | | |
| 80meV/atom | 8.90% | 16.35% | 60.92% | 34.17% | | 1.8%[7] | 0.36%[28] |

**Table 2|Percent of structures with energy above the hull per atom lower than the given standard among these batches and some other models.**

Moreover, we chose three materials ($Ca_2SnO_4$, $LiMg_6$, and $MgSc_2O_4$) from the generated structures for further investigation (as shown in Fig. 5). We utilized DS-PAW[34] for structural relaxation calculations and band structure assessments. Importantly, all three materials were successfully optimized via DFT calculations. Subsequently, we conducted phonon structure calculations for these selected materials, and all the materials demonstrated structural stability. Among these three compounds, $MgSc_2O_4$ and $LiMg_6$ are being reported for the first time, highlighting the novelty of this research. These compounds are notably challenging to obtain through simple elemental substitution. Furthermore, among these three materials, $MgSc_2O_4$ and

LiMg$_6$ were first reported, but they are difficult to obtain by simple elemental substitution. Furthermore, Cheng et al.[35], who used the PCCD to discover of Magnesium-Aluminum alloys, demonstrated that PCCD can generate structures effectively. This finding not only validates the use of PCCD in the discovery and design of new materials but also opens new avenues for future research in material science.

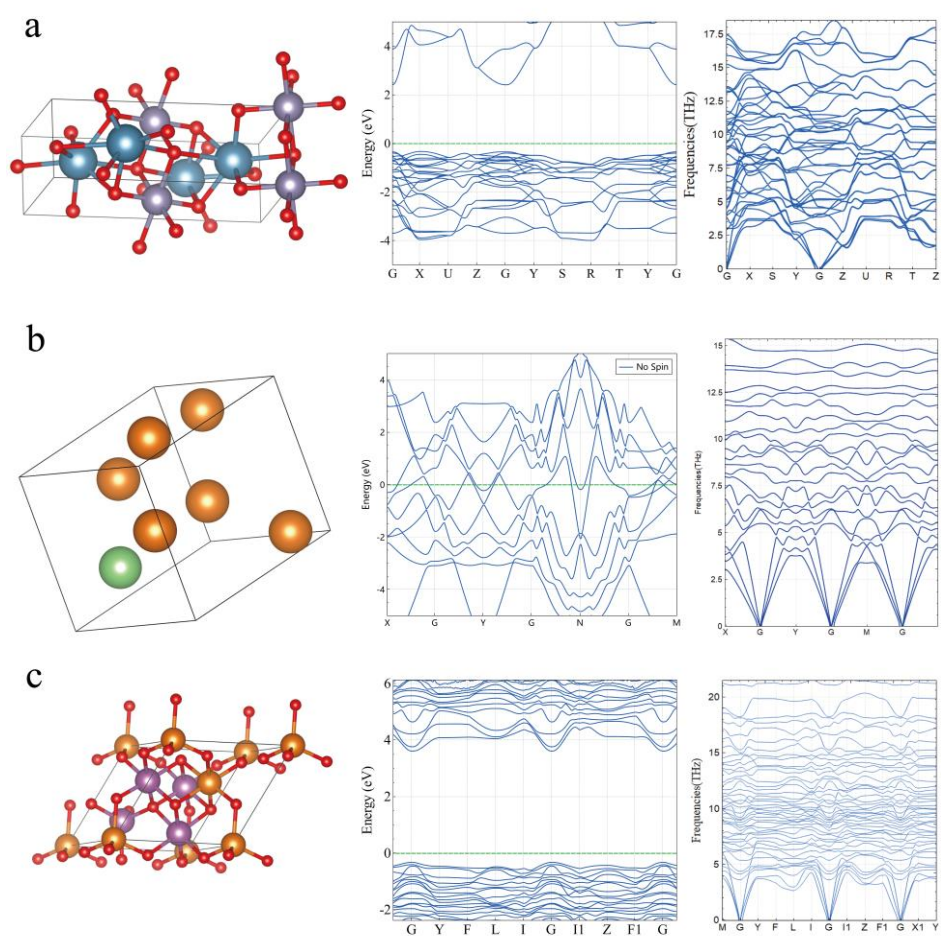

**Fig. 5|Graphical depiction of the structures and their DFT calculations.** Crystal representation and band and phonon band structures of Ca$_2$SnO$_4$ (a), LiMg$_6$ (b) and MgSc$_2$O$_4$(c).

## Methods

At the core of our approach is the utilization of a diffusion model as the foundational model, as illustrated in Fig. 6. We leverage U-Net[36] as the backbone of PCCD, a well-established architecture frequently employed for tasks such as classification and segmentation tasks[37,38].

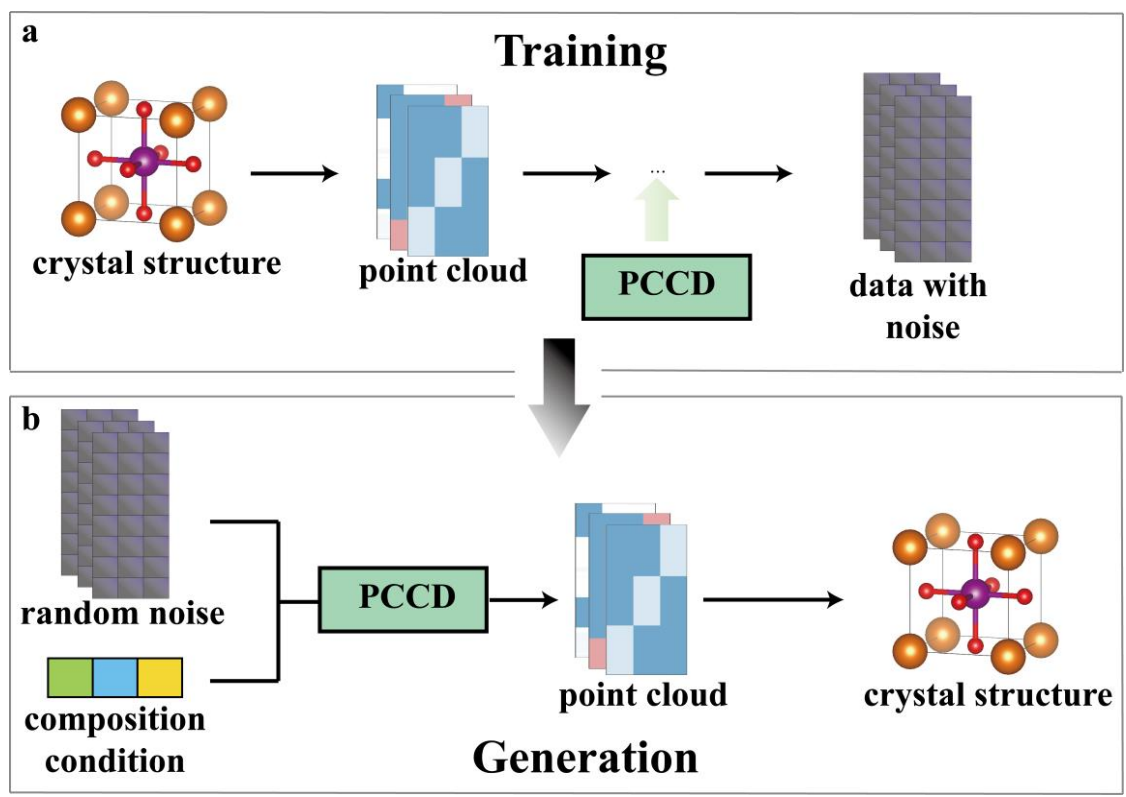

**Fig. 6|Sketch map of the PCCD. Here, we highlight the data flow of the framework. a.** Training phase process with data manipulation section. First, crystals are transformed to a point cloud data type, followed by the addition of noise to the data, which enables the PCCD to perform observation and learning; **b.** Generation phase with retrieval data operation. The method starts by feeding the PCCD random data and composition conditions and then passes to the data extraction and finishing with generating structures.

Two main methods are commonly used to represent a 3D object: voxels and point clouds. Voxel-based representation is thorough but resource intensive. In contrast, point clouds are more efficient than sparse matrices and reduce resource usage. Some prior works have claimed to use point clouds[7], but they essentially used individual points to lower computational costs. Our approach treats point clouds and lattice constants as three-channel entities akin to RGB in the computer vision (CV) field. We then employ clustering to determine the position, element composition and lattice.

Drawing from the notable achievements of diffusion models in the field of computer vision, we are motivated to extend their application to the generation of crystal structures. In this paradigm, we envision each crystal structure as akin to a patch in an image. To explore this innovative approach further, we integrate the point cloud representation technique with the power of diffusion models within the PCCD. This fusion of methods is designed to leverage the inherent advantages of both approaches. The diffusion model, renowned for its ability to capture intricate dependencies in data, holds promise for encoding the structural nuances of crystal formations. Moreover, the use of point cloud data representation, akin to a cloud of 3D points, serves to describe atomic positions and their attributes efficiently. By combining these two methodologies, we seek to harness their collective potential to revolutionize the generation and understanding of crystal structures.

## DATA PREPROCESSING

Our material data were sourced from the Materials Project (MP)[2]. In this extensive database, our selection process targeted structures with ternary, binary, or monadic compositions that feature a maximum of 16 atom sites. This thorough filtering yielded a comprehensive dataset comprising 52,028 distinct materials. This dataset encompasses a wealth of information, including the POSCAR file, band gap, magnetism, crystal system, magnetic ordering, etc., for each of these structures. However, for model efficiency, we opted to narrow our focus to the band gap and magnetic ordering as the primary control variables. This decision, in conjunction with our use of the POSCAR files as training data, was made to streamline and lighten the model while ensuring the retention of essential variables for our specific research objectives.

As mentioned previously, we initially gathered various properties and POSCAR files of each crystal before training. The primary objective revolves around transforming the POSCAR data into a three-channel format, encompassing atom positions, element information, and lattice constants, as illustrated in Fig. 7. Each of these channels comprises 128 items, effectively representing each structure as a 3×128×3 (C×W×H) matrix. The first channel is dedicated to atomic site information, where we distribute 128 points within the space. It is essential to clarify that the positions here are relative coordinates akin to those in the POSCAR file. The lattice vectors have not been determined at this stage. In essence, we use 128 items or several sets of data at this point in the process. To determine the absolute positions of these points, it is necessary to multiply them by the three lattice vectors obtained after processing the third channel. The data in the second channel correspond one-to-one with those in the first channel. Prior to generating or training samples, we input up to three elements. Each item in this channel contains three data values, which represent the likelihood of these three elements being associated with each atom. The data in the third channel do not correspond one-to-one with those in the first two channels. In fact, we want to obtain only six parameters $\alpha, \beta, \gamma, a, b, c$ from here, which can be converted to three vectors. To match the shape before, we expand them to 128 items by copying. In theory, after training, two distinct groups of data are generated. We can then obtain three vectors by employing clustering techniques, determining the means of each group, and performing calculations.

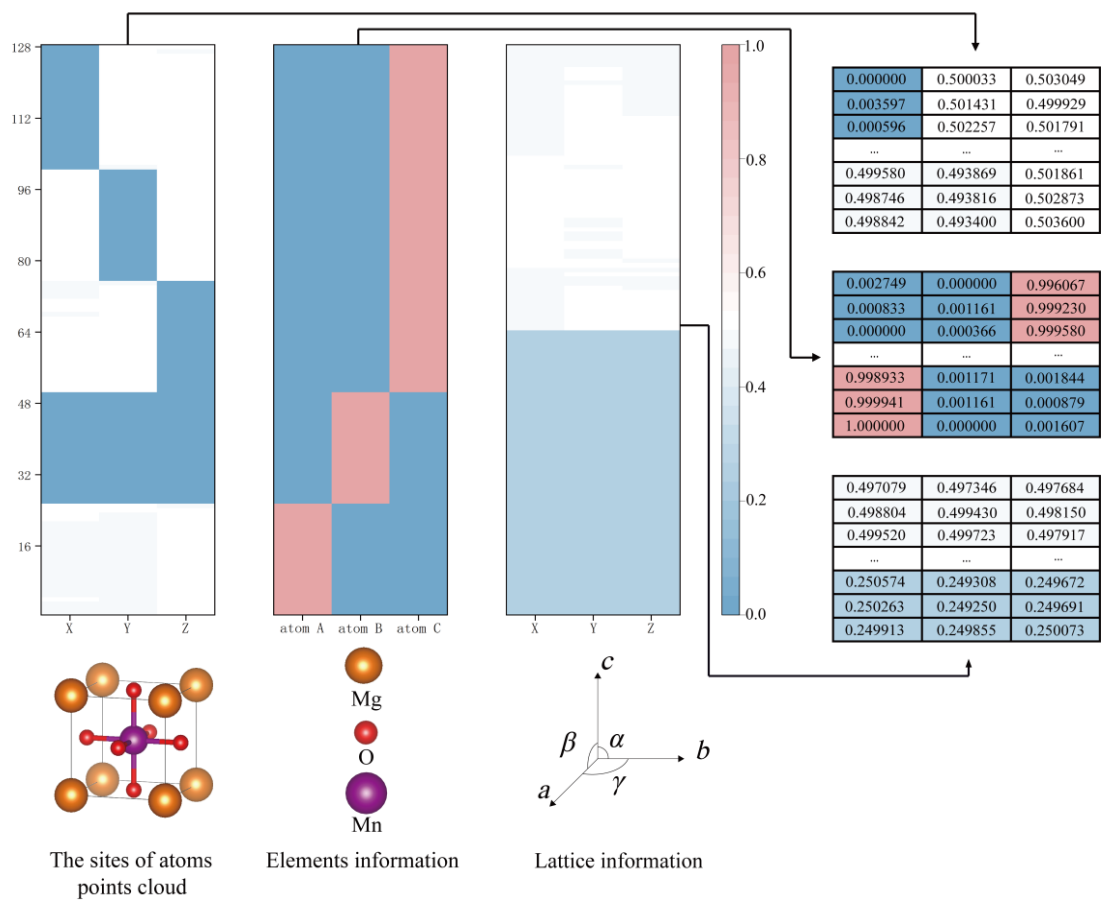

**Fig. 7|Data format for the framework (e.g., MgMnO$_3$).** The first channel represents the position, the second channel represents the element information, and the third channel determines the lattice constants.

# GENERATION MODEL

Our generation model is based on the diffusion model, which is essentially a parameterized Markov chain. It is trained using variational inference to produce samples that closely match the data distribution after finite time[20].

The diffusion model comprises two distinct processes, the training process and the generation process, often referred to as the sampling process, as illustrated in Fig. 8a. These processes work in tandem to enable the generation of data samples that align with the underlying distribution of the training data. The training process can be briefly described as a procedure in which noise is progressively introduced to the data and the model endeavors to meet the characteristics of this noise addition. In contrast, the sampling process involves the gradual application of the trained model to denoise pure noise data. These data, in essence, are treated as source data with superimposed noise, and the model works to refine and clarify them.

The training process begins with $x_0$ and gradually adds noise $\varepsilon_1, \varepsilon_2, \cdots, \varepsilon_{T-1}, \varepsilon_T$ to $x_0$, resulting in $x_1, x_2, \cdots, x_{T-1}, x_T$. Assuming that $x_0 \sim q(x_0)$ and the noise $\varepsilon_t$ follow a normal distribution, then, for $t \geq 1$:

$$q(x_{1:T}|x_0) = \prod_{t=1}^{T} q(x_t|x_{t-1}) \qquad q(x_t|x_{t-1}) = \mathcal{N}\left(x_t; \sqrt{1-\beta_t}x_{t-1}, \beta_t I\right) \qquad (1)$$

We follow the definition of J. Ho et al.[20] Here, we define a constant variance schedule $\beta_1, \cdots, \beta_T$, where β increases as t increases. According to reparameterization, equation (1) can also be expressed as:

$$x_t = \sqrt{1-\beta_t}x_{t-1} + \sqrt{\beta_t}\,\epsilon \qquad (2)$$

where $\epsilon \sim \mathcal{N}(0,1)$. We can obtain $x_t$ through the probability method from $x_{t-1}$. For simplicity, we define $\alpha_t = 1 - \beta_t$, $\overline{\alpha_t} = \prod_{s=1}^{t} \alpha_s$ and $\overline{\beta_t} = \prod_{s=1}^{t} \beta_s$. By applying equation (2) recursively, we can obtain that at any time t:

$$x_t = \sqrt{\overline{\alpha_t}}x_0 + \sqrt{1-\overline{\alpha_t}}\epsilon, \qquad q(x_t|x_0) = \mathcal{N}(x_0; \sqrt{\overline{\alpha_t}}x_0, (1-\overline{\alpha_t})I) \qquad (3)$$

and the reverse process begins with $p(x_T) = \mathcal{N}(x_T, \mathbf{0}, \mathbf{I})$; this process denoises gradually as $p_\theta(x_0) = \int p_\theta(x_{0:T}) dx_{1:T}$. In the reverse process $p_\theta$, we know the variance of every step but do not know the means $\boldsymbol{\mu_\theta}$.

$$p_\theta(x_{0:T}) = p(x_T) \prod_{t=1}^{T} p_\theta(x_{t-1}|x_t) \quad p_\theta(x_t|x_{t-1}) = \mathcal{N}(x_t; \mu_\theta(x_t, t), \beta_t I) \quad (4)$$

Therefore, we need to know what $\boldsymbol{\mu_\theta}$ is. It can be derived that[20,39]:

$$\mu_\theta(x_t, t) = \frac{1}{\sqrt{\alpha_t}} \left( x_t - \frac{\beta_t}{\sqrt{1-\bar{\alpha}_t}} \epsilon_\theta(x_t, t) \right) \quad (5)$$

After the parameterization (5), for any $t \in [1, T]$:

$$x_{t-1} = \frac{1}{\sqrt{\alpha_t}} \left( x_t - \frac{\beta_t}{\sqrt{1-\bar{\alpha}_t}} \epsilon_\theta(x_t, t) \right) + \sigma_t z \quad (6)$$

$\epsilon_\theta$ is the model that needs to be trained. This means that we can obtain $x_{t-1}$ from $x_t$ by $\epsilon_\theta$.

In a one-step noise addition process (Fig. 8.b①), the noise is composed of random numbers following a normal distribution. The mean and variance of this noise depend on time $t$ and the preceding data $x_{(t-1)}$. Simultaneously, the PCCD actively learns the characteristics of this noise. During each iteration, the model receives data with noise, and its primary task is to predict the most recent noise addition. Consequently, we end up with two types of noise: one generated from a probabilistic approach and the other predicted by our deep learning model. By comparing these two noise sources, we can calculate a loss, which serves as feedback to the model, facilitating its adjustment and improvement. This iterative process continues until the model effectively learns to reproduce the noise characteristics, achieving accurate denoising.

In the noise-removal process, as shown in Fig. 8b②, we only have data with noise, and our objective is to estimate and separate the noise from the data. In this way, we can separate the current noise and the previous data. At the macro level, this is a disorderly to orderly process.

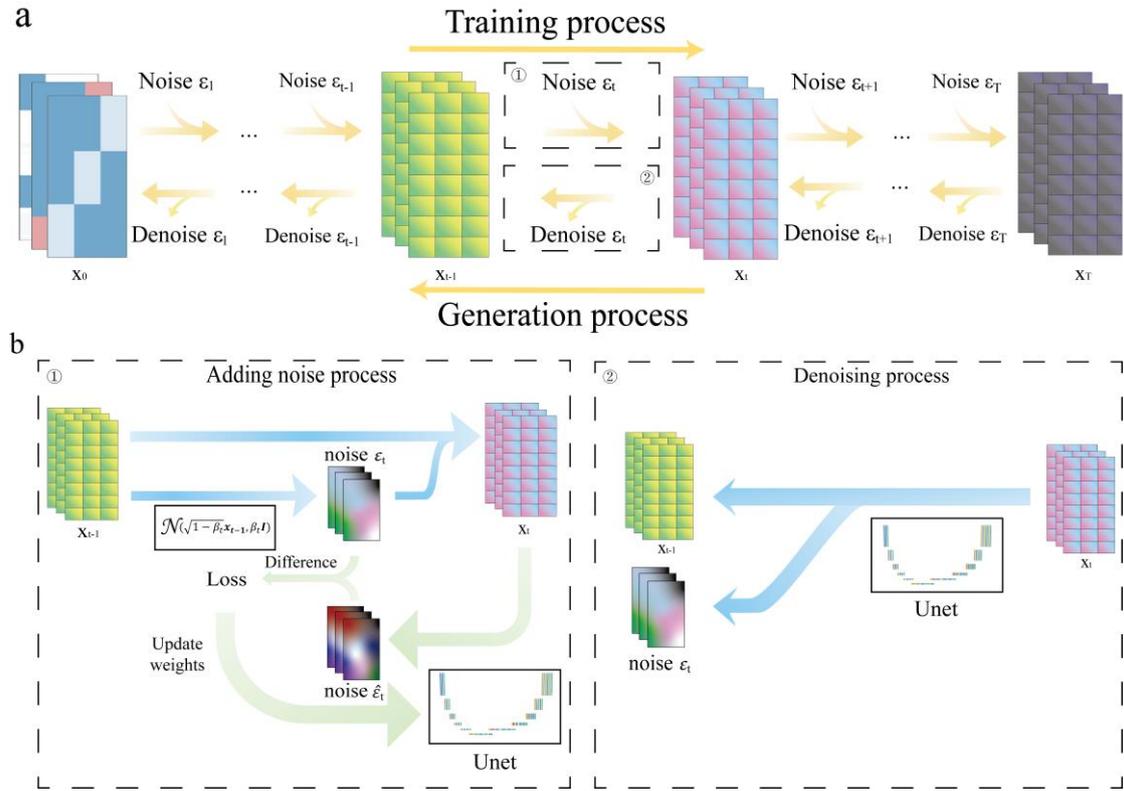

**Fig. 8|Schematic depiction of the PCCD architecture with generation and training processes. a.** The data flow of the training process and generation process. **b①.** A step of the training process of the model corresponds to part ① in a. **b②.** A step of the denoising process corresponds to part ② in a.

In this context, we employ a U-Net model (Fig. S3) to predict and separate noise from the data. U-Net, initially introduced in 2015[36], is a well-established model in the CV field that was notably acclaimed for its exceptional performance in image segmentation tasks. Our U-Net model is configured with five sets of upsampling and downsampling layers. To enhance its capacity to capture intrinsic data correlations, we incorporated intra-data correlation. This augmentation allows the model to effectively learn and predict noise, contributing to the denoising process.

## Conclusion and future work

We introduced a framework employing the denoising diffusion probabilistic model (DDPM) and point cloud representation for crystal structure generation. This versatile framework enables the generation of crystal structures composed of fewer than three elements and featuring up to 16 atom sites by specifying the elemental composition. To assess the framework's validity, we successfully reconstructed a batch of structures randomly sampled from the training dataset, conffirming its reliability. Furthermore, we applied this framework to generate a batch of structures comprising rare earth elements, alkaline earth elements, transition metal elements and oxygen as an illustrative example. For the three batches of crystals generated by the PCCD, the percentages of structures with $E_{hull}$/atom less than 0.25 eV/atom were 39.44%, 61.85% and 66.67%, respectively, and these with $E_{hull}$/atom less than 80 meV/atom were 8.90%, 16.35% and 34.17%, respectively. Structures with some special components are more abundant. In addition, the stabilities of several structures have been confirmed through phonon structure analysis (e.g., $Ca_2SnO4$, $LiMg_6$, and $MgSc_2O_4$). Consequently, we demonstrated the efficacy of utilizing DDPM and point cloud representations in crystal structure generation, which was validated by DFT high-throughput calculations. This framework serves as a foundational step, offering potential for further enhancement and the development of larger models for inverse crystal design. Furthermore, this approach serves to expand the database of crystals.

# Experimental procedures

## Model training progress

In PCCD, we employ the U-Net for noise prediction (Fig S3). It primarily encompasses four up-sampling progresses and for down-sampling progresses, with each progresses comprising multiple convolutional layers and self-attention layers.

As depicted in formula (3), for each step *t* during the training progress, we can calculate $x_t$, while the noise $\epsilon_t \sim \mathcal{N}(0,1)$ is given. The objective of the U-Net is to estimate the noise term $\epsilon_t$ given $x_t$. For the loss of U-Net, we utilize Mean Absolute Error (MAE, formula (S4)) to quantify the discrepancy between output of U-Net and $\epsilon_t$. For more training detail can be seen in Supplemental Information.

## Resource availability

## Lead contact

Further information and requests for resources should be directed to and will be fulfilled by the lead contact, Shibing Chu(c@ujs.edu.cn).

## Materials availability

This study did not generate new unique reagents.

## Data availability

The crystal datasets used and analyzed during the current study are available in the Materials Project (https://next-gen.materialsproject.org/). Relevant data that support the key findings of this study are available within the article, the Supplemental Information, the batch1.csv batch2.csv and batch3.csv files. All raw data generated in the current study are available from the corresponding author upon request.

## Code availability

The code for this PCCD is available at https://github.com/lzhelin/CrystalDiffusion and https://zenodo.org/records/10570395

## Acknowledgements

This work gratefully acknowledges the financial support of the funding No. 4111190003 from Jiangsu University (JSU) and the National Natural Science Foundation of China (No. 11904137, 12074150 and 12174157). We gratefully acknowledge HZWTECH for providing computation facilities.

## Author contributions

Z.L.: conceptualization, method, software, investigation, formal analysis, model validation, writing original draft; R.M: editing, model validation and investigation; R.J, G.H and J.S: DFT calculation guide; S.C and Y.C: conceptualization, writing review, funding acquisition, resources and supervision.

## Declaration of interests

The authors declare no competing interests.

## References<sub>Uncategorized References</sub>

## Supplemental Information

### Data expression example

Taking MgMnO$_3$ as an example (Fig. S1), the data is generated by our diffusion model. For the first channel, obviously, 128 data points can be classified into five categories: (0, 0.5, 0.5), (0.5, 0, 0.5), (0.5, 0.5, 0), (0, 0, 0) and (0.5, 0.5, 0.5). The clustering method we use is Density-Based Spatial Clustering of Applications with Noise (DBSCAN) due to the raison of undefined number of groups in our data. However, here we may know: one atom is at the position of (0, 0, 0), one atom is at the center of the crystal body, and three atoms are at the centers of the crystal faces. However, we have not been given more information on elements and lattice. It can be a carbon system or a Ca-Ti-O system, and can also be a triclinic system or a hexagonal system. At the second channel, it gives the element information for the 128 points one by one. In Fig. S1, it can be seen that there are three groups of data, further categorize the five categories in the first channel into three by one-to-one correspondence", we change like follow "the first data channel is divided to five classes of data while the second is divided to three classes. However, these two channels (first and second) are aligned according to three classes of data (second channel) due to index correspondence (see how classes are aligned in figure S1). and they further categorize the five categories in the first channel into three by one-to-one correspondence. So far, the elements and the relative coordinate in unit cell of every atom can be confirmed. Before training, the elements information have been given by inputting a list (Mg, Mn, O), which assign in the second channel: (1,0,0), (0,1,0), and (0,0,1) mean Mg, Mn, O, separately. But the shape of lattice or crystal system is still unknown, and it can also be a cubic system or a trigonal system. Ignoring the first two channels, while processing the third channel, we deal with them directly. It can be aggregated into two categories theoretically which are the lenghth $a, b, c$ and angle $\alpha, \beta, \gamma$ of lattice. During training and data preprocess, all samples have the same template for the third channel. Hence, we neglect the use of additional clustering algorithm, only perform calculations average for every column of the front half ($\alpha$, $\beta$ and $\gamma$) and back half (a, b and c). Here, the data we get approximately are (0.50, 0.50, 0.50), (0.25, 0.25, 0.25) (All numbers below 1 is because before training, lattice data have been normalized by dividing 15Å and the angle used radian system and divided by $2\pi$ ). Then, the three vectors of the lattice can be calculated as follows.

$$\vec{a} = a(1,0,0) \tag{S1}$$

$$\vec{b} = b\big((cos\gamma, sin\gamma, 0)\big) \tag{S2}$$

$$\vec{c} = c(cos\beta, cos\alpha - \frac{cos\beta cos\gamma}{sin\gamma}, \frac{\sqrt{1+2cos\alpha cos\beta cos\gamma - cos^2\alpha - cos^2\beta - cos^2\gamma}}{sin\gamma}) \tag{S3}$$

The result is (3.75, 0, 0), (0, 3.75, 0), (0, 0, 3.75), and further determine that it's a cubic system with side length of 3.75Å. Integrate the above all, we summarize:
1. It's a cubic system with side length of 3.75Å.
2. The formula of this structure is MgMnO$_3$.

3. For every unit cell, there will be an oxygen atom on each face, a magnesium atom on each corner, and a manganese atom at the center.

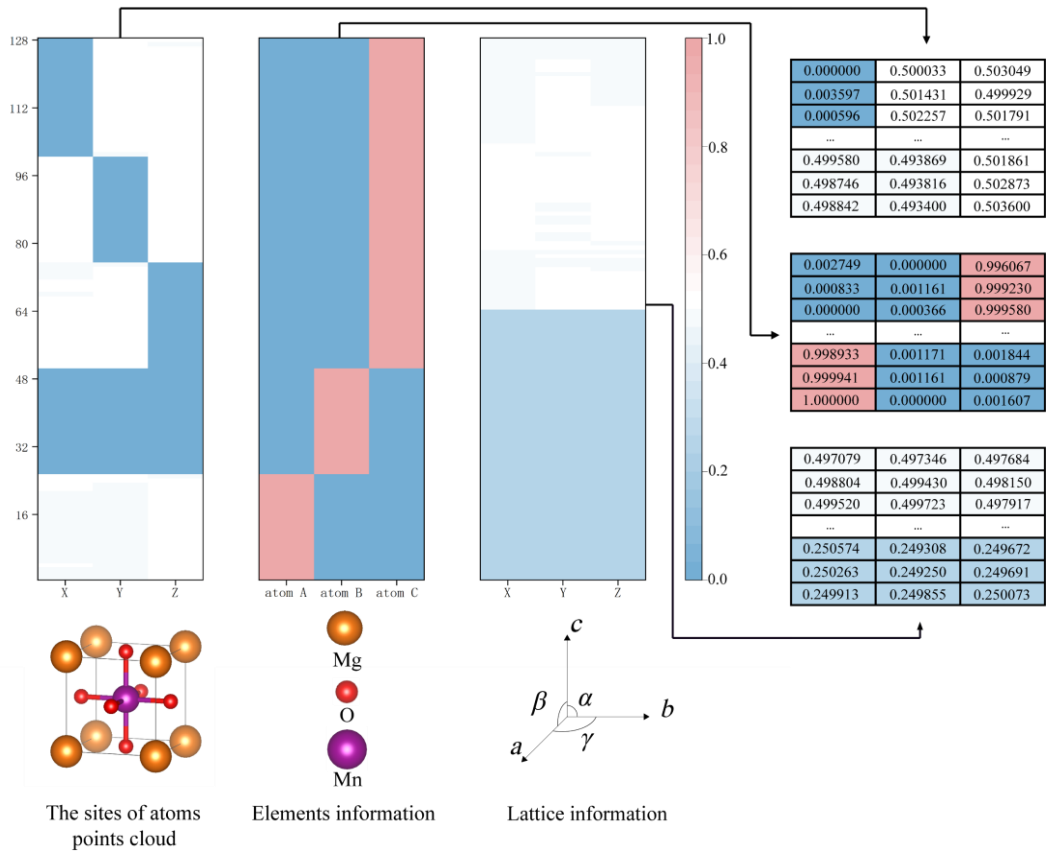

**Fig S1. Data expression example.**

## Generation processes example

Also taking MgMnO$_3$ as an example(Fig. S2), different color of points means different atoms (the second channel of data) and ignoring the third channel for better visualization. The whole inference process need 1000 steps. It's the last step for training and also the first step for generation while t=999, while generating, the data given here is random numbers obey a normal distribution. It can be seen that the points in this step are disorganized. with t getting closer and closer to 0. Points with the same color gradualy come together. From t=200, there are already rudiments of clustering. While t=0, it can be clear that there are five groups, and it means that this structure has these five atoms, including positions and elements.

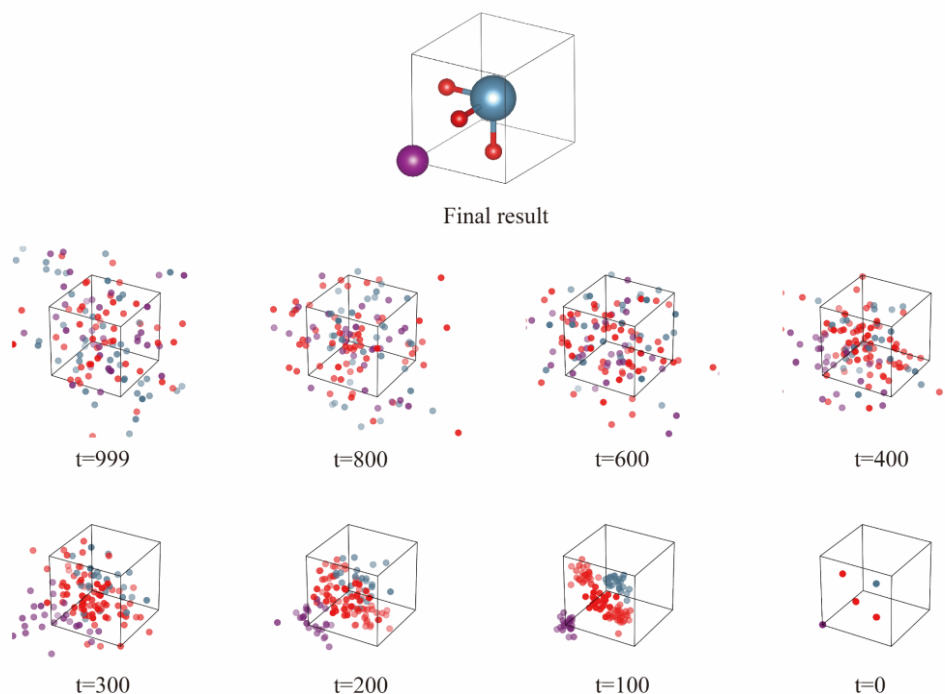

**Fig S2. Generation processes example.**

## DFT configuration

The structures were optimized by Density Functional Theory (DFT) that were carried out with Vienna ab initio simulation package (VASP). The Perdew–Burke–Ernzerhof (PBE) of the generalized gradient approximation (GGA) was used for exchange–correlation functional. The kinetic energy cutoff was set to be 520 eV for the electronic wavefunction having a plane wave basis set which was obtained using the projector augmented-wave method. The Monkhorst–pack k-mesh grids was selected by vaspkit.

## Model details

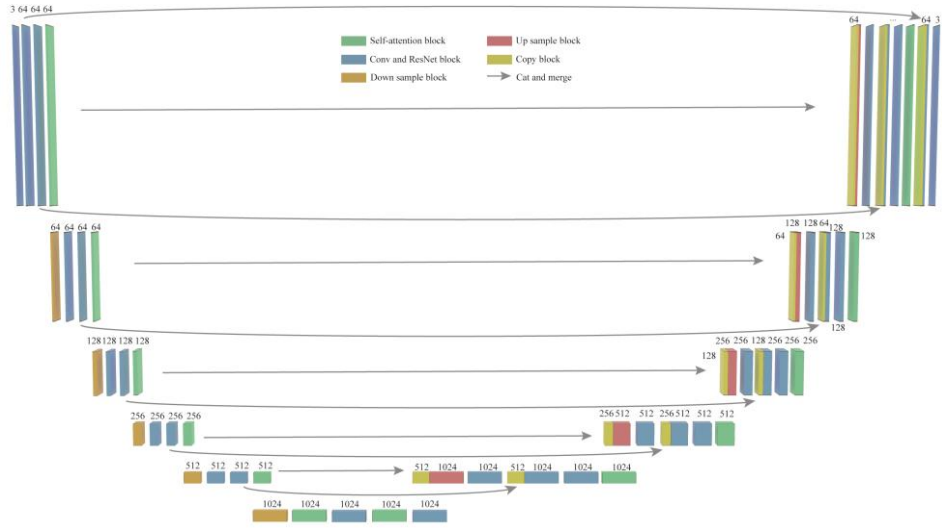

**Fig S3. U-Net model used in diffusion.**

## Training details

Hyperparameters while training are as Table. S1.

| Hyperparameter | Value |
|---|---|
| Optimizer | Adam ($\beta_1 = 0.9$, $\beta_2 = 0.99$) |
| Learning rate | 0.0001 |
| Batch size | 128 |
| Training steps | 500 |
| Prediction target | $\epsilon$ |
| Sampling timesteps | 1000 |
| Diffusion noise schedule | cosine |

**Table S1. Hyperparameters for training.**

While training, we use Mean Absolute Error (MAE) to calculate loss (Fig. S4). The loss function is as follow.

$$Loss = \mathbb{E}_{t,\mathbf{x_0},\epsilon}(\|\epsilon - \epsilon_\theta(\sqrt{\bar{\alpha}_t}x_0 + \sqrt{1-\bar{\alpha}_t}\epsilon, t)\|) \qquad (S4)$$

Here, $t$ is timestep, $x_0$ is the original data (training data) without noise, $\epsilon$ is random matrices with the same shape of $x_0$ following normal distribution.

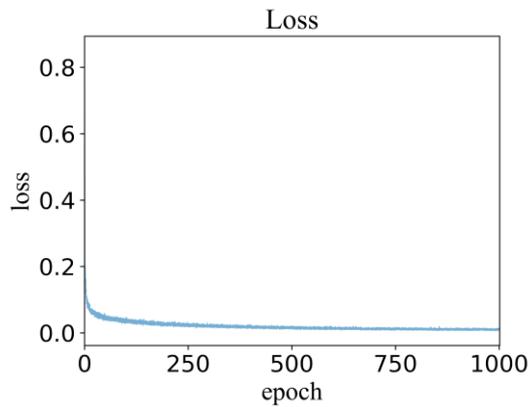

**Fig. S4 Loss plot.**

## Generation Data

All 3 batches structures generated and their energy above hull can be seen in <u>batch1.csv, batch2.csv and batch3.csv.</u>